\documentclass[10pt,letterpaper]{article}

\usepackage[margin=1in]{geometry}

\title{Video-to-Model: Automatic Modeling of Deformable Linear Objects}
\author{Akshun Sharma, Kimia Forghani, and Yancy Diaz-Mercado\\[0.5em]
\small Department of Mechanical Engineering, University of Maryland\\
\small College Park, Maryland, USA\\
\small \texttt{\{asharm41, kimiaf, yancy\}@umd.edu}
\thanks{This manuscript has been submitted to the 2027 American Control Conference (ACC) and is currently under review.}
\thanks{The authors used OpenAI ChatGPT Codex (GPT-5.6 Sol) to assist with code development. All outputs were independently verified by the authors, who assume full responsibility for the article's content.}}
\date{}

\usepackage{amsmath}
\usepackage{amssymb}
\usepackage[dvipsnames]{xcolor}
\usepackage{makecell}
\usepackage{graphicx}
\usepackage{multirow}
\usepackage{array}
\usepackage{hhline}
\definecolor{algcomment}{RGB}{30,150,30}

\begin{document}

\maketitle

\begin{abstract}
This paper presents a video-to-model framework for automatically modeling the motion of a deformable suture thread from an input video. We utilize a recently developed CBF--CLF--QP numerical model that simplifies the characterization of deformable string motion through the selection of a small number of parameters. A perception module first localizes and tracks the thread in video, producing an ordered sequence of thread nodes. The observed thread motion is then processed by a spatio-temporal CNN network that estimates the effective parameters of a structured CBF--CLF--QP model. These parameters are used to simulate the thread under a user-defined needle velocity input. Experiments using unseen thread configurations and motion demonstrate that the framework can reliably reconstruct the thread behavior from video, automatically configure the structured model, and reproduce the expected thread motion with low tracking error. The proposed approach reduces the need for manual parameter tuning and provides a step toward automatic video-based modeling of deformable linear objects.
\end{abstract}

\section{Introduction}

Deformable linear objects (DLOs) are encountered in applications such as robotic suturing~\cite{jackson2015suture}, cable routing~\cite{monguzzi2023cable},
and wire-harness assembly~\cite{jiang2011wire}, where predicting their behavior can support reliable robotic perception, planning, and control. However, their flexibility gives them many degrees of freedom, making them difficult to model. Unlike a rigid object, whose motion can be described
using a small number of state variables, DLOs can be represented in different ways depending on the modeling approach~\cite{arriola2020modeling}. For example, a common discrete representation approximates a DLO as a
sequence of connected nodes whose state is defined by their positions and velocities~\cite{chen2025deform}. 

In this work, we focus specifically on surgical suture threads, which are highly flexible DLOs whose
motion is governed by their geometry, material properties, and interaction with the surrounding environment. Furthermore, variations in bending, stretching, and stiffness can produce different thread responses to the same input.~\cite{arriola2020modeling}.
Consequently, accurately modeling thread behavior remains challenging.

\begin{figure}
    \centering
    \includegraphics[width=0.75\linewidth]{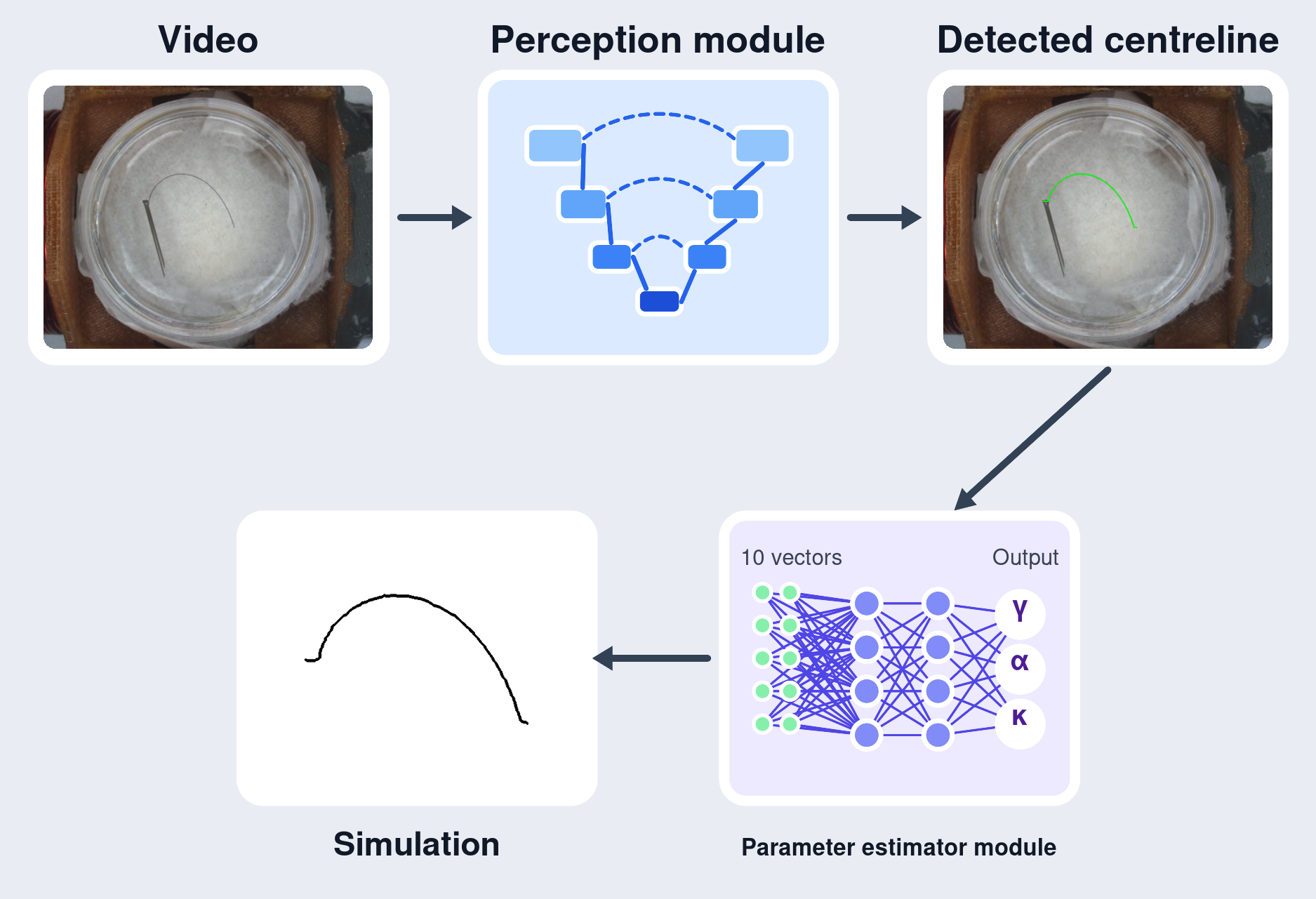}
    \caption{Overview of the proposed video-to-model framework for automatic thread-model configuration. An input video is fed into a perception module to detect the centerline of the DLO. This object is then fed into a parameter estimator module which enables a CBF--CLF--QP simulation of the object.}
    \label{fig1}
\end{figure}

Accurately modeling the thread is particularly important in robot-assisted suturing, where a
robot must anticipate the thread’s response to needle motion. Conventional physics-based approaches describe a thread using masses, elastic forces, damping forces, and contact models~\cite{qi2017virtual,xu2018realtime}. However, more modeled elements increase computational cost, while uncertain material properties and unmodeled dynamics can reduce prediction accuracy. 

Another formulation introduced in~\cite{forghani2025suture} represents the thread using a collection of ordered nodes governed by a
control-barrier-function and control-Lyapunov-function quadratic program
(CBF--CLF--QP). Instead of explicitly calculating all mechanical forces on the thread, the
model computes node velocities that satisfy structural constraints. Connectivity CBFs prevent neighboring nodes from
separating beyond a safe threshold, while a CLF drives the thread
to its rest state to model stiffness. The QP objective ensures that the resulting motion is limited to what is
necessary to satisfy the CBF and CLF constraints, each corresponding to specific aspects of thread behavior.

Although this framework provides a structured way of modeling thread dynamics, its parameters must be selected so that the simulated thread matches the observed response to the same needle motion. This motivates an approach that identifies model parameters directly from thread observations rather than through manual tuning.
This work addresses this gap by developing a video-to-model framework that automatically configures the CBF--CLF--QP model from observed thread motion. Given a video of a moving thread, the proposed method observes its motion and estimates the parameters of the framework
introduced in~\cite{forghani2025suture}, producing a thread-specific model whose simulated behavior reproduces the observed motion.

The main contributions of this work are: \textit{(i)} a U-Net~\cite{ronneberger2015unet} based perception module that localizes the
thread in video frames and converts it into a
temporally consistent trajectory of ordered nodes; \textit{(ii)} a spatio-temporal Convolutional Neural Network (CNN) module that uses the tracked response of the thread to needle motion to estimate model parameters and automatically configure the CBF--CLF--QP framework.

The rest of the paper is organized as follows.
Section~\ref{sec:background} introduces the CBF--CLF--QP formulation. Section~\ref{sec:problem_statement} defines the problem statement.
Section~\ref{sec:methods} presents the proposed localization, tracking, model-configuration, and simulation pipeline.
Section~\ref{sec:results} presents the experimental results, and
Section~\ref{sec:conclusion} concludes the paper and discusses future work.

\section{Background}
\label{sec:background}

This section summarizes the CBF--CLF--QP formulation introduced 
in~\cite{forghani2025suture}, which models the suture thread by
constraining the velocities of a collection of discrete nodes rather than by
explicitly computing mechanical forces.

\subsection{Discrete Suture-Thread Model}
\label{subsec:discrete_thread_model}
The thread is discretized into \(N\) planar nodes, where
\(
\mathbf{x}_i=
\begin{bmatrix}
x_i & y_i
\end{bmatrix}^{\mathsf T}
\in\mathbb{R}^2
\)
denotes the position of node \(i\), and \(\mathbf{x}_0\) denotes the lead node corresponding to the needle. Each node is modeled as a single integrator,
\begin{equation}
    \dot{\mathbf{x}}_i=\mathbf{u}_i,
    \qquad i=1,\ldots,N,
\end{equation}
while the needle satisfies \(\dot{\mathbf{x}}_0=\mathbf{u}_0\), where \(\mathbf{u}_0\) is the user-provided needle velocity. The thread states and velocities are stacked as
\(
\mathbf{x}=
\begin{bmatrix}
\mathbf{x}_1^{\mathsf T} & \cdots & \mathbf{x}_N^{\mathsf T}
\end{bmatrix}^{\mathsf T},
\qquad
\mathbf{u}=
\begin{bmatrix}
\mathbf{u}_1^{\mathsf T} & \cdots & \mathbf{u}_N^{\mathsf T}
\end{bmatrix}^{\mathsf T}.
\)
The QP determines \(\mathbf{u}\) subject to the geometric constraints governing thread behavior, and the node positions are updated using forward Euler method.

\subsection{Primer on CBFs and CLFs}
\label{subsec:cbf_clf_background}
The following section is largely adapted from \cite{forghani2025suture}. Consider a control-affine dynamical system
\begin{equation}
\dot{\mathbf{x}}=\mathbf{f}(\mathbf{x})+\mathbf{g}(\mathbf{x})\mathbf{u}.
\end{equation}
where \(\mathbf{x}\in\mathbb{R}^{d}\) is the state and
\(\mathbf{u}\in\mathbb{R}^{m}\) is the control input.

A CBF is used to prevent a system from entering
an unsafe state. Let \(h(\mathbf{x})\) be a continuously differentiable
function defining the safe set
\begin{align}
    \mathcal{C}
    &=
    \left\{
        \mathbf{x}\in\mathbb{R}^{d}
        \mid
        h(\mathbf{x})\geq0
    \right\},
\end{align}

In this work, the thread is considered to violate connectivity when the
distance between two adjacent nodes exceeds the maximum allowable separation
\(\Delta>0\). This condition is encoded by the connectivity CBF:
\begin{equation}
    h^{\mathrm{con}}_i(\mathbf{x})
    =
    \frac{K_\mathrm{con}}{2}
    \left(
        \Delta^2
        -
        \left\|
            \mathbf{x}_i-\mathbf{x}_{i-1}
        \right\|_2^2
    \right),
    \qquad i=1,\ldots,N.
    \label{eq:CBF}
\end{equation}
for $K_\mathrm{con}>0$.

A sufficient condition for keeping the safe set
\(\mathcal{C}\) forward invariant~\cite{ames2019control}, such
that \(h_i^{\mathrm{con}}(\mathbf{x}(t))\geq0\) for all time, is
\begin{equation}
    \dot{h}_i^{\mathrm{con}}(\mathbf{x},\mathbf{u})
    \geq
    -\gamma h_i^{\mathrm{con}}(\mathbf{x}),
    \qquad
    \gamma>0,
    \label{eq:gamma}
\end{equation}
where \(\gamma\) determines how strongly the system responds as the
connectivity constraint approaches its boundary. Intuitively, the choice of $\gamma$ controls the stretchiness of the thread.

A CLF is used to model thread stiffness by driving the thread toward a
preferred rest configuration. It is defined as
\begin{align}
    V_{i-1}^{\mathrm{stiff}}(\mathbf{x})
    =
    \frac{K_\mathrm{stiff}}{2}
    \left(
        \left\|
            \mathbf{x}_i-\mathbf{x}_{i-2}
        \right\|_2^2
        -
        \delta_i^2
    \right)^2,
    &&& i=2,\ldots,N,
\end{align}
for $K_\mathrm{stiff}>0$, where \(\delta_i>0\) is the desired separation between nodes \(i\) and
\(i-2\). The CLF is nonnegative and becomes zero when this desired separation
is achieved. The condition
\begin{equation}
    \dot{V}_{i-1}^{\mathrm{stiff}}(\mathbf{x},\mathbf{u})
    \leq
    -\alpha V_{i-1}^{\mathrm{stiff}}(\mathbf{x}),
    \qquad
    \alpha>0,
    \label{eq:alpha}
\end{equation}
drives the thread toward its rest configuration, where \(\alpha\) determines
the rate at which the CLF decreases.

Because the CBF and CLF derivative conditions are affine in the control
input, they can be imposed as constraints in a quadratic program. The CBF--CLF--QP can be written as
\begin{equation}
\begin{aligned}
    \mathbf{u}^{*}_t
    =
    \arg\min_{\mathbf{u}_t,\,s_t}
    \quad&
    \frac{1}{2}
    \left\|
        \mathbf{u}_t-\mathbf{v}^{\mathrm{des}}_t
    \right\|_{\mathbf{Q}}^{2}
    +
    \frac{p}{2}s_t^{2}
    \\
    \text{subject to}\quad&
    \mathbf{v}^{\mathrm{des}}_t
    =
    \kappa\mathbf{u}^{*}_{t-1},
    \\
    &
\dot{h}^{\mathrm{con}}_i(\mathbf{x}_t,\mathbf{u}_t)
\geq
-\gamma h^{\mathrm{con}}_i(\mathbf{x}_t),
\\
&
\dot{V}^{\mathrm{stiff}}_{i-1}(\mathbf{x}_t,\mathbf{u}_t)
\leq
-\alpha V^{\mathrm{stiff}}_{i-1}(\mathbf{x}_t)+s_t,
\\
&
s_t\geq0.
\end{aligned}
\end{equation}
Here, \(\mathbf{u_t}^{*}\) is the modified input, \(s_t\) relaxes the CLF constraint when needed to preserve feasibility, and \(p\) penalizes this relaxation.
$\mathbf{v}^{\mathrm{des}}_t$ is generally set to zero, modeling that the thread does not move unless the connectivity constraint is being violated by the needle pulling on it. When the thread moves in a liquid, it may experience some drift when $\mathbf{u}_0$ becomes zero. The residual drift is modeled by
\begin{equation}
    \mathbf{v}^{\mathrm{des}}_{i,t}
    =
    \kappa\mathbf{u}_{i,t-1},
    \qquad
    0\leq\kappa\leq1.
    \label{eq:kappa}
\end{equation}
where \(\mathbf{u}_{i,t-1}\) is the node velocity produced at the preceding time step and \(\kappa\) is the
velocity-retention factor. A larger value of \(\kappa\) produces more
persistent drift, whereas a smaller value causes a drag-like effect leading to the retained motion decay more rapidly.

The QP attempts to follow the desired motion as closely as
possible while modifying it whenever necessary to enforce connectivity and stiffness. Further details on the construction of $h$ and $V$ can be found in~\cite{forghani2025suture}. Different choices of \(\gamma\), \(\alpha\), and \(\kappa\) produce different simulated thread behaviors. This paper focuses on automatically determining these parameters from observed thread motion, as in the
following section.

\section{Problem Statement}
\label{sec:problem_statement}

Let
\begin{equation}
    \mathcal{V}
    =
    \left\{
        (I_t,\tau_t)
    \right\}_{t=0}^{T-1}
\end{equation}
denote a timestamped monocular video of a thread undergoing planar motion in
response to a moving needle in a fluid, where
\(I_t\in\mathbb{R}^{H\times W\times 3}\) is the image recorded at time
\(\tau_t\). 
The problem consists of estimating the thread
configuration over time from video and identifying the effective CBF--CLF--QP parameters in \eqref{eq:real_theta} below so that the simulated motion matches the observed thread behavior.

\subsection{Thread State Observation from Video}
\label{subsec:thread_localization_problem}

At time \(\tau_t\), the thread configuration is represented by the planar curve
\(
    \mathbf{c}_t:[0,L_t]\rightarrow\mathbb{R}^{2},
\)
where \(L_t\) is the curve length. The complete thread curve can be represented as
\begin{equation}
    \mathcal{C}_t
    =
    \left\{
        \mathbf{c}_t(s)
        \mid
        s\in[0,L_t]
    \right\}
    \subset\mathbb{R}^{2}.
\end{equation}
where \(s\) is the distance along the thread and \(\mathbf{c}_t(0)\) is the needle-attached endpoint.

The recorded frame is modeled as:
\begin{equation}
    I_t
    =
    \mathcal{H}
    \left(
        \mathcal{C}_t,
        \boldsymbol{\eta}_t
    \right)
    +
    \boldsymbol{\epsilon}_t,
\end{equation}
where \(\mathcal{H}\) is the unknown observation operator that maps the scene into the recorded image plane, \(\boldsymbol{\eta}_t\) captures visual effects such as illumination, reflections, occlusions, and motion blur, and \(\boldsymbol{\epsilon}_t\) is measurement noise.

The first task is to estimate the time-varying thread curve from the video:
\begin{equation}
    \mathcal{V}
    =
    \left\{
        (I_t,\tau_t)
    \right\}_{t=0}^{T-1}
    \longmapsto
    \left\{
        \widetilde{\mathcal{C}}_t
    \right\}_{t=0}^{T-1},
\end{equation}
where \(\widetilde{\mathcal{C}}_t\) is the estimated thread curve  at time \(\tau_t\).

For compatibility with the CBF--CLF--QP model in
Section~\ref{sec:background}, the curve is discretized into \(N\)
ordered nodes:
\begin{equation}
    \widetilde{\mathbf{X}}_t
    =
    \begin{bmatrix}
        \widetilde{\mathbf{x}}_{1,t} &
        \widetilde{\mathbf{x}}_{2,t} &
        \cdots &
        \widetilde{\mathbf{x}}_{N,t}
    \end{bmatrix}
    \in\mathbb{R}^{2\times N},
    \label{eq:thread_nodes}
\end{equation}
where
\(\widetilde{\mathbf{x}}_{i,t}\in\widetilde{\mathcal{C}}_t\) for
\(i=1,\ldots,N\).
The nodes are ordered from the needle-attached endpoint to
the opposite endpoint. The estimated needle attachment point is
\(\widetilde{\mathbf{x}}_{0,t}
\). 

The output of the thread-observation problem is therefore:
\begin{equation}
    \mathcal{O}_{\mathcal{V}}
    =
    \left\{
        \widetilde{\mathbf{X}}_t,
        \widetilde{\mathbf{x}}_{0,t},
        \tau_t
    \right\}_{t=0}^{T-1},
    \label{eq:observation}
\end{equation}
which provides the trajectory used for subsequent parameter identification.

\subsection{Video-Based Dynamical Parameter Identification}
\label{subsec:parameter_identification_problem}

Let the vectorized thread state and observed needle state at time \(\tau_t\) be
\begin{equation}
    \widetilde{\mathbf{z}}_t
    =
    \operatorname{vec}\!\left(\widetilde{\mathbf{X}}_t\right)
    \in\mathbb{R}^{2N},
    \qquad
    \widetilde{\mathbf{u}}_t
    =
    \begin{bmatrix}
        \widetilde{\mathbf{x}}_{0,t}^{\mathsf T} &
        \dot{\widetilde{\mathbf{x}}}_{0,t}^{\mathsf T}
    \end{bmatrix}^{\mathsf T}
    \in\mathbb{R}^{4}
\end{equation}
The observed thread dynamics are modeled as:
\begin{equation}
\widetilde{\mathbf{z}}_{t+1}
=
\mathcal{F}_{\mathrm{CBF\text{-}CLF\text{-}QP}}
\left(
\widetilde{\mathbf{z}}_t,
\widetilde{\mathbf{u}}_t;
\boldsymbol{\theta}
\right)
+
\mathbf{w}_t,
\end{equation}
where \(\mathcal{F}_{\mathrm{CBF\text{-}CLF\text{-}QP}}\) denotes the CBF--CLF--QP model, \(\mathbf{w}_t\) represents unmodeled dynamics, and
\begin{equation}
    \boldsymbol{\theta}
    =
    \begin{bmatrix}
        \gamma & \alpha & \kappa
    \end{bmatrix}^{\mathsf T}
    \in\Theta
    \label{eq:real_theta}
\end{equation}
is the unknown effective parameter vector. Here, \(\gamma\), \(\alpha\), and
\(\kappa\) govern the connectivity response \eqref{eq:gamma}, stiffness response
\eqref{eq:alpha}, and velocity retention \eqref{eq:kappa}, respectively.
Unlike geometric quantities such as thread length obtained directly from the observed curve, these
parameters are inferred from the thread motion and assumed constant over each video.

The subsequent problem is to determine a parameter-identification mapping
\begin{equation}
    g_{\boldsymbol{\omega}}:
    \mathcal{O}_{\mathcal{V}}
    \longrightarrow
    \Theta
\end{equation}
such that
\begin{equation}
    \widehat{\boldsymbol{\theta}}
    =
    g_{\boldsymbol{\omega}}
    \left(
        \mathcal{O}_{\mathcal{V}}
    \right)
    =
    \begin{bmatrix}
        \widehat{\gamma} &
        \widehat{\alpha} &
        \widehat{\kappa}
    \end{bmatrix}^{\mathsf T}.
    \label{eq:theta_hat}
\end{equation}
Here, \(g_{\boldsymbol{\omega}}\) denotes a learning-based model with trainable weights \(\boldsymbol{\omega}\).

Given \(\widehat{\boldsymbol{\theta}}\), the simulation is initialized from the
first observed state and propagated using observed needle motion:
\begin{equation}
\begin{aligned}
    \widehat{\mathbf{z}}_0 
    &=
    \widetilde{\mathbf{z}}_0, \\
    \widehat{\mathbf{z}}_{t+1}
    &=
    \mathcal{F}_{\mathrm{CBF\text{-}CLF\text{-}QP}}
    \left(
        \widehat{\mathbf{z}}_t,
        \widetilde{\mathbf{u}}_t;
        \widehat{\boldsymbol{\theta}}
    \right),
    \qquad t=0,\ldots,T-2.
    \label{eq:dynamical_propagation}
\end{aligned}
\end{equation}

The simulated and observed trajectories are compared using the mean node
tracking error
\begin{equation}
    \mathcal{J}_{\mathrm{track}}
    \left(
        \boldsymbol{\theta}
    \right)
    =
    \frac{1}{(T-1)N}
    \sum_{t=1}^{T-1}
    \sum_{i=1}^{N}
    \left\|
        \widehat{\mathbf{x}}_{i,t}
        \left(
            \widehat{\boldsymbol{\theta}}
        \right)
        -
        \widetilde{\mathbf{z}}_{i,t}
    \right\|_2
\end{equation}
\(\widehat{\mathbf{x}}_{i,t}
(\widehat{\boldsymbol{\theta}})\) denotes the position
of node \(i\) at time \(\tau_t\) generated by the CBF--CLF--QP model using the parameter vector \(\widehat{\boldsymbol{\theta}}\).

The parameter vector that best reproduces the observed thread motion
is defined by the inverse dynamical problem
\begin{equation}
    \boldsymbol{\theta}^{*}
    =
    \arg\min_{\boldsymbol{\theta}\in\Theta}
    \mathcal{J}_{\mathrm{track}}
    \left(
        \boldsymbol{\theta}
    \right).
\end{equation}
and the learned mapping is trained such that
\begin{equation}
    \widehat{\boldsymbol{\theta}}
    =
    g_{\boldsymbol{\omega}}
    \left(
        \mathcal{O}_{\mathcal{V}}
    \right)
    \approx
    \boldsymbol{\theta}^{*}.
\end{equation}
The predicted configuration
\(\widehat{\boldsymbol{\theta}}\) is subsequently used to instantiate the
CBF--CLF--QP model of the observed thread.

\section{Methods}
\label{sec:methods}

\begin{figure}
    \centering
    \includegraphics[width=0.75\linewidth,
    height=0.4\textheight,
    keepaspectratio]{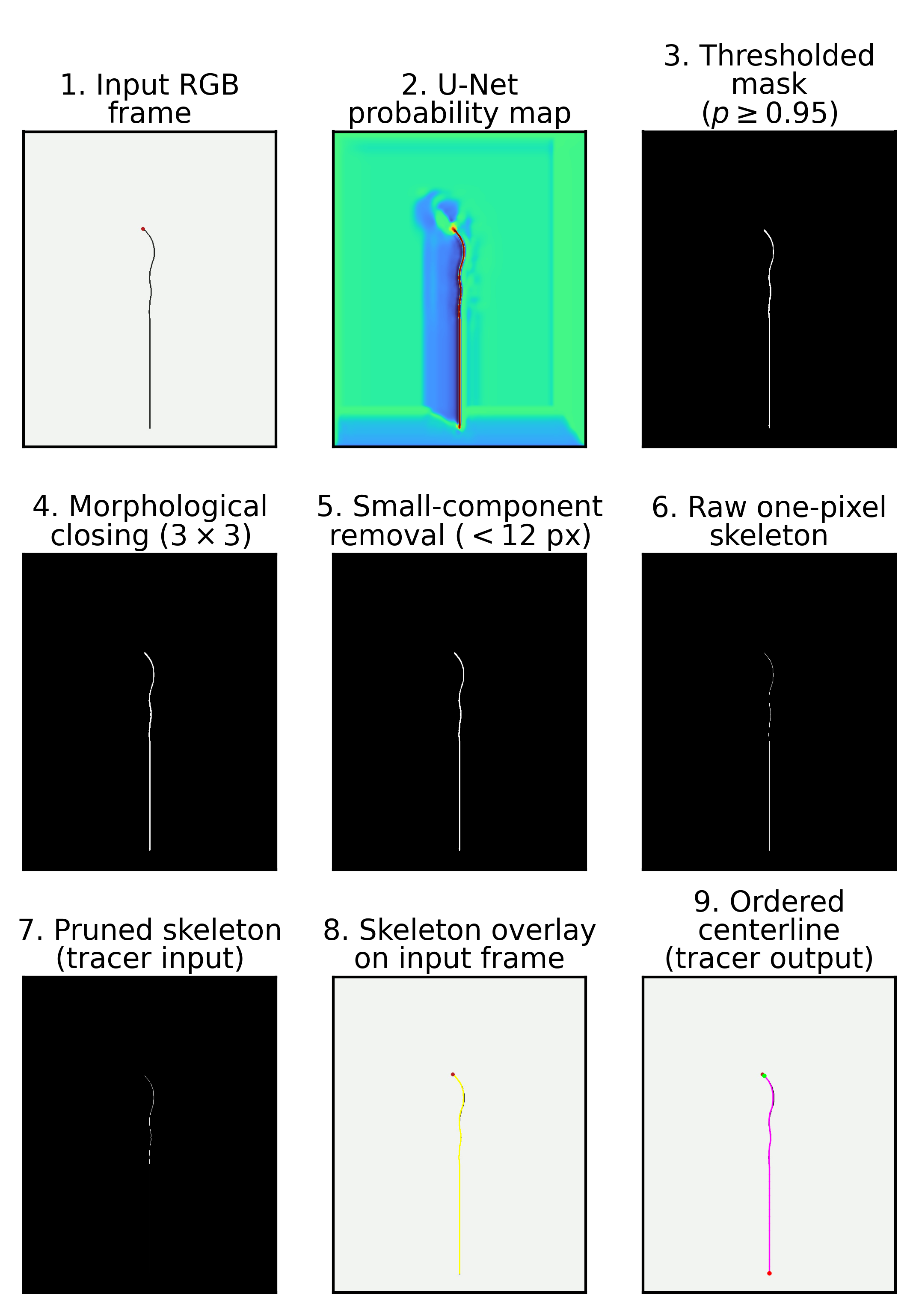}
    \caption{Thread-localization and tracking pipeline. An input RGB frame is processed by the U-Net to obtain a thread-probability map, which is thresholded and cleaned using morphological closing and small-component removal. The resulting mask is skeletonized, pruned, and traced to obtain an ordered thread centerline. The green and red markers indicate the two centerline endpoints.}
    \label{fig:placeholder}
\end{figure}

\subsection{Thread Localization and Tracking Module}
\label{subsec:thread_localization_method}

\subsubsection{U-Net Segmentation}

Each RGB frame is normalized to the interval \([0,1]\) and passed through a four-level U-Net to produce a single-channel thread probability map:
\begin{equation}
    P_t
    =
    \sigma
    \left(
        S_{\boldsymbol{\phi}}
        \left(
            \frac{I_t}{255}
        \right)
    \right),
\end{equation}
where \(S_{\boldsymbol{\phi}}\) denotes the U-Net with trainable parameters
\(\boldsymbol{\phi}\), \(\sigma(\cdot)\) is the sigmoid function, and
\(P_t(p)\in[0,1]\) represents the probability that pixel \(p\) belongs to the
thread.

The U-Net contains four encoder levels with {8, 16, 32, and 64} channels,
followed by a 128-channel bottleneck. Each level uses two convolutional
layers, group normalization, and SiLU activation functions. 

The segmentation dataset is generated from simulated thread trajectories, rendered as paired RGB frames and ground-truth binary masks. Background appearance, illumination, thread color, thread width,
image noise, and needle appearance are randomized during rendering. The dataset contains ten videos with 120 frames per video. Six complete
videos are assigned to training, two to validation, and two to testing.

Training uses \(256\times256\) image crops. Most crops are centered near the
thread because thread pixels occupied only a small fraction of the complete
image. Horizontal and vertical reflections, brightness and contrast changes,
gamma adjustment, Gaussian noise, and image blur are used as augmentation.

The training objective combines three losses:
\begin{equation}
    \mathcal{L}_{\mathrm{seg}}
    =
    0.35\mathcal{L}_{\mathrm{BCE}}
    +
    0.35\mathcal{L}_{\mathrm{Dice}}
    +
    0.30\mathcal{L}_{\mathrm{clDice}}.
\end{equation}
The weighted binary cross-entropy (BCE) loss term  handles the imbalance between thread
and background pixels, the Dice loss term measures overall mask overlap, and the
centerline Dice (clDice) loss term encourages the predicted thread to remain connected.

Training was performed on a workstation equipped with an AMD Ryzen~7 4800H
processor, \(16\,\mathrm{GB}\) of RAM, and an NVIDIA GeForce RTX~3050 Laptop
GPU with \(4\,\mathrm{GB}\) of video memory. The network was trained using the
AdamW optimizer~\cite{loshchilov2019decoupled} with a batch size of eight,
weight decay of \(10^{-4}\), a cosine learning-rate schedule, gradient
clipping, and mixed-precision GPU training. After each epoch, several
probability thresholds were evaluated on the validation videos. The checkpoint
and threshold producing the highest validation Dice score were retained.

\subsubsection{Mask Post-Processing}

To reduce frame-to-frame fluctuations, the current probability map is
combined with the filtered prediction from the preceding frame:
$\overline{P}_t = 0.7P_t + 0.3\overline{P}_{t-1}$.
A binary mask is then produced as $M_t(p)=\mathbb{I}\!\left[\overline{P}_t(p)\geq 0.95\right]$.

A \(3\times3\) morphological closing operation~\cite{serra1982image} is applied to connect small
breaks in the mask, after which connected components containing fewer than
12 pixels are removed. The cleaned mask is then skeletonized to obtain a
one-pixel-wide representation of the thread. Short branches caused by
segmentation noise are pruned before curve tracking.

\subsubsection{Curve Tracking}

The thread skeleton is represented as a graph containing endpoints,
junctions, and the paths connecting them. At ordinary positions, the thread
is traced by following neighboring skeleton pixels. At self-crossings, the
continuation is selected using the local curve direction together with the
thread direction estimated in the preceding frame. This temporal information
reduced sudden changes in curve ordering at crossings.

The needle endpoint is localized separately using the centroid of the colored needle
marker. Of the two thread endpoints, the endpoint closest to the needle is
selected as the beginning of the ordered curve. This ensured that the
direction of the traced thread remained consistent across frames. If the
standard graph traversal failed, an Euler-path traversal~\cite{bondy2008graph} was a
fallback.

The curve is sampled at approximately equal arc-length intervals to
produce \(N=30\) nodes in pixel coordinates. Missing detections were interpolated from neighboring valid frames, and a
Savitzky--Golay filter~\cite{savitzky1964smoothing} was applied to reduce high-frequency tracking noise.
Finally, the pixel coordinates were converted into physical coordinates using
the camera calibration into~\eqref{eq:thread_nodes}. The output of this module is~\eqref{eq:observation}. The resulting thread and needle trajectories form the input to the spatio--temporal CNN module.

\subsection{Spatio-Temporal CNN Module}
\label{subsec:parameter_identification_method}

\subsubsection{Temporal Alignment and Velocity Estimation}

The video frames are sampled at
intervals of \(0.048\,\mathrm{s}\) to match the time step used to generate the simulated trajectories.

Node velocity is estimated by calculating rate of change of their respective positions in consecutive frames over time.
\begin{equation}
    \widetilde{\mathbf{v}}_{i,t}
    =
    \frac{
        \widetilde{\mathbf{x}}_{i,t}
        -
        \widetilde{\mathbf{x}}_{i,t-1}
    }{
        \Delta t
    },
    \qquad
    i=1,\ldots,N.
\end{equation}
The observed needle velocity can also be calculated in a similar way.

\subsubsection{Spatial--Temporal Feature Construction}

A twelve-dimensional feature vector is constructed for every node and
every frame. The thread length is \(L\) (e.g., \(200\,\mathrm{mm}\)). For \(N\) discrete thread nodes, the nominal
spacing is \(\Delta=L/(N-1)\).

Since the inputs~\eqref{eq:feature_vector} to the network are normalized,
the velocity features are scaled independently for each trajectory. The
normalization scale \(v_s\) is calculated as the \(95\)th percentile of the
combined thread-node and needle speed magnitudes, with a small positive lower
bound \(\epsilon_v\) to prevent division by zero. All thread and needle
velocity components are divided by this common scale. 
To preserve information about the absolute motion speed, the quantity
\(s_v=\log(1+v_s/v_{\mathrm{ref}})/5\), where
\(v_{\mathrm{ref}}=1\,\mathrm{mm/s}\), is included as an additional
speed-context feature for every node and frame.

The distance between the needle endpoint and the first node is
\(d_{1,t}=\lVert\widetilde{\mathbf{x}}_{1,t}
-\widetilde{\mathbf{x}}_{0,t}\rVert_2\). For the remaining nodes,
\(d_{i,t}=\lVert\widetilde{\mathbf{x}}_{i,t}
-\widetilde{\mathbf{x}}_{i-1,t}\rVert_2\), for \(i=2,\ldots,N\).

The normalized CBF connectivity function \eqref{eq:CBF} is defined
\begin{equation}
    h_{i,t}^{\mathrm{con}}
    =
    \frac{K_\mathrm{con}}{2}
    \left[
        1-
        \left(
            \frac{d_{i,t}}{\Delta}
        \right)^2
    \right]
\end{equation}

Local bending is represented by the change in direction between two adjacent nodes. The direction of the segment from node \(i\) to node
\(i+1\) is
\begin{equation}
    \mathbf{a}_{i,t}
    =
    \frac{
        \widetilde{\mathbf{x}}_{i+1,t}
        -
        \widetilde{\mathbf{x}}_{i,t}
    }{
        \left\|
            \widetilde{\mathbf{x}}_{i+1,t}
            -
            \widetilde{\mathbf{x}}_{i,t}
        \right\|_2+\epsilon
    },
\end{equation}
where \(\epsilon>0\) is a small constant used to avoid division by zero.
The local bending feature at node \(i\) is then
\begin{equation}
    c_{i,t}
    =
    \left\|
        \mathbf{a}_{i,t}
        -
        \mathbf{a}_{i-1,t}
    \right\|_2.
\end{equation}

Thus, the complete feature vector can be represented as
\begin{equation}
\mathbf{f}(i,t)=
\left[
\begin{gathered}
\tfrac{\widetilde{x}_{i,t}-\widetilde{x}_{0,t}}{L},\,
\tfrac{\widetilde{y}_{i,t}-\widetilde{y}_{0,t}}{L},\,
\tfrac{\widetilde{v}_{x,i,t}}{v_s},\,
\tfrac{\widetilde{v}_{y,i,t}}{v_s},\,
\tfrac{d_{i,t}}{\Delta},\,
c_{i,t}
\\[2pt]
h_{i,t}^{\mathrm{con}},\,
\tfrac{i-1}{N-1},\,
\tfrac{\dot{\widetilde{x}}_{0,t}}{v_s},\,
\tfrac{\dot{\widetilde{y}}_{0,t}}{v_s},\,
s_v,\,
V_{i,t}^{\mathrm{stiff}}
\end{gathered}
\right]^{\mathsf T}
\label{eq:feature_vector}
\end{equation}
which includes:
(1) horizontal node position relative to the needle; (2) vertical node position relative to the needle; (3) normalized horizontal node velocity; (4) normalized vertical node velocity; (5) normalized distance from the preceding node or needle; (6) local thread curvature; (7) normalized connectivity-CBF; (8) normalized position along the thread; (9) normalized horizontal needle velocity; (10) normalized vertical needle velocity; (11) the trajectory speed-context feature; and (12) the normalized stiffness-CLF.

Stacking the features over \(T\) frames and \(N\) nodes produces
\begin{equation}
    \mathbf{F}
    =
    \left[
        \mathbf{f}_{i,t}
    \right]
    \in
    \mathbb{R}^{T\times N\times 12}
\end{equation}

\subsubsection{Spatial Node Encoder}

The spatio-temporal CNN receives an input batch (\(B\)) of shape
\(B\times T\times N\times 12\).
Each node feature vector is first mapped into a 64-dimensional representation:
\begin{equation}
    \mathbf{e}^{(0)}_{i,t}
    =
    \operatorname{MLP}_{\mathrm{point}}
    \left(
        \mathbf{f}_{i,t}
    \right)
    \in\mathbb{R}^{64}
\end{equation}
Here, MLP refers to multi-layer perceptron. The node encoder consists of two linear layers with SiLU activation and layer normalization (LN).

The representations of all \(N\) nodes in frame \(t\) are collected as
\(\mathbf{E}^{(0)}_t
=
[\mathbf{e}^{(0)}_{1,t},\ldots,\mathbf{e}^{(0)}_{N,t}]^{\mathsf T}
\in\mathbb{R}^{N\times64}\), where \(\mathbf{E}^{(\ell)}_t\) denotes the
node representations after convolutional block \(\ell\).

Three residual one-dimensional convolutional blocks are then applied along
the ordered node dimension:
\begin{equation}
    \mathbf{E}^{(\ell+1)}_t
    =
    \operatorname{LN}
    \left(
        \mathbf{E}^{(\ell)}_t
        +
        \operatorname{SiLU}
        \left(
            \operatorname{Conv1D}_{7}
            \left(
                \mathbf{E}^{(\ell)}_t
            \right)
        \right)
    \right),
\end{equation}
where \(\operatorname{Conv1D}_{7}\) denotes a one-dimensional convolution with
kernel size seven and \(\ell=0,1,2\). These convolutions allow every node
representation to gather information from neighboring nodes along
the thread.

\subsubsection{Spatial Attention and Frame Representation}

Not every part of the thread is equally informative for identifying the parameters in \eqref{eq:real_theta}. A spatial attention weight is therefore computed for
each node:
\begin{equation}
    a_{i,t}
    =
    \frac{
        \exp\left(
            \mathbf{w}_a^{\mathsf T}\mathbf{e}_{i,t}
        \right)
    }{
        \sum_{j=1}^{N}
        \exp\left(
            \mathbf{w}_a^{\mathsf T}\mathbf{e}_{j,t}
        \right)
    }
\end{equation}
where \(\mathbf{w}_a\in\mathbb{R}^{64}\) is a trainable attention vector. The weights satisfy
\(\sum_{i=1}^{N}a_{i,t}=1\).
The spatially pooled representation of frame \(t\) is
\begin{equation}
    \mathbf{p}_t
    =
    \sum_{i=1}^{N}
    a_{i,t}\mathbf{e}_{i,t}.
\end{equation}

Spatial normalization reduces information about the absolute magnitude of
the raw physical features. For this reason, the mean, standard deviation,
minimum, and maximum of each of the twelve input features are also calculated
over the \(N\) nodes:
\begin{equation}
\begin{aligned}
\mathbf{m}_t
=
\operatorname{concat}\Big(
&\operatorname{mean}_{i}(\mathbf{f}_{i,t}),
\operatorname{std}_{i}(\mathbf{f}_{i,t}),\\
&\operatorname{min}_{i}(\mathbf{f}_{i,t}),
\operatorname{max}_{i}(\mathbf{f}_{i,t})
\Big)
\in\mathbb{R}^{48}
\end{aligned}
\end{equation}

The attended representation and raw feature statistics are fused into one
64-dimensional frame embedding:
\begin{equation}
    \mathbf{g}_t
    =
    \operatorname{MLP}_{\mathrm{frame}}
    \left(
        \begin{bmatrix}
            \mathbf{p}_t\\
            \mathbf{m}_t
        \end{bmatrix}
    \right)
    \in\mathbb{R}^{64}
\end{equation}

\subsubsection{Temporal Encoding and Attention}

The sequence of frame embeddings are processed by a two-layer gated recurrent
unit (GRU)~\cite{cho2014learning}:
\begin{equation}
    \mathbf{h}_{1:T}
    =
    \operatorname{GRU}
    \left(
        \mathbf{g}_{1:T}
    \right),
    \qquad
    \mathbf{h}_t\in\mathbb{R}^{64}.
\end{equation}
The GRU allows the representation at time \(t\) to depend on both the current
thread shape and the preceding motion history.

Four temporal attention heads are used to identify informative portions of
the observed motion. For attention head \(r\), the temporal weights are
\begin{equation}
    \beta_{r,t}
    =
    \frac{
        \exp\left(
            \mathbf{w}_r^{\mathsf T}
            \tanh\left(
                \mathbf{W}_r\mathbf{h}_t
            \right)
        \right)
    }{
        \sum_{\tau=0}^{T-1}
        \exp\left(
            \mathbf{w}_r^{\mathsf T}
            \tanh\left(
                \mathbf{W}_r\mathbf{h}_{\tau}
            \right)
        \right)
    }.
\end{equation}
The attended temporal representation is
\begin{equation}
    \mathbf{a}_r
    =
    \sum_{t=0}^{T-1}
    \beta_{r,t}\mathbf{h}_t.
\end{equation}
Using multiple heads allows the network to attend to different phases of the
motion, such as stretching, bending, reversal, release, and velocity decay.

The mean, standard deviation, minimum, and maximum of the frame embeddings
are also computed over time. These global temporal statistics are
concatenated with the four attended representations to form a sequence-level
embedding.

\subsubsection{Parameter Prediction and Scaling}

The sequence-level embedding is passed through a two-layer prediction head
with SiLU activation and dropout. A sigmoid output produces three normalized
values in the interval \([0,1]\), corresponding to \(\gamma\), \(\alpha\), and
\(\kappa\). These values are subsequently mapped to the parameter ranges
\(\gamma\in[1,100]\), \(\alpha\in[1,100]\), and
\(\kappa\in[0,1]\). Since \(\gamma\) and \(\alpha\) span relatively
large multiplicative ranges, they are encoded and decoded on a logarithmic
scale. The velocity-retention parameter \(\kappa\) is scaled linearly.

\subsubsection{Virtual-Material Dataset}

Training data for parameter identification is generated by treating each
combination of
\begin{equation}
    \boldsymbol{\theta}^{(m)}
    =
    \begin{bmatrix}
        \gamma^{(m)} &
        \alpha^{(m)} &
        \kappa^{(m)}
    \end{bmatrix}^{\mathsf T}
\end{equation}
as one virtual material \(m\). 200 virtual materials are generated
using random sampling over the parameter ranges.

\begin{figure}
    \centering
    \includegraphics[width=1\linewidth]{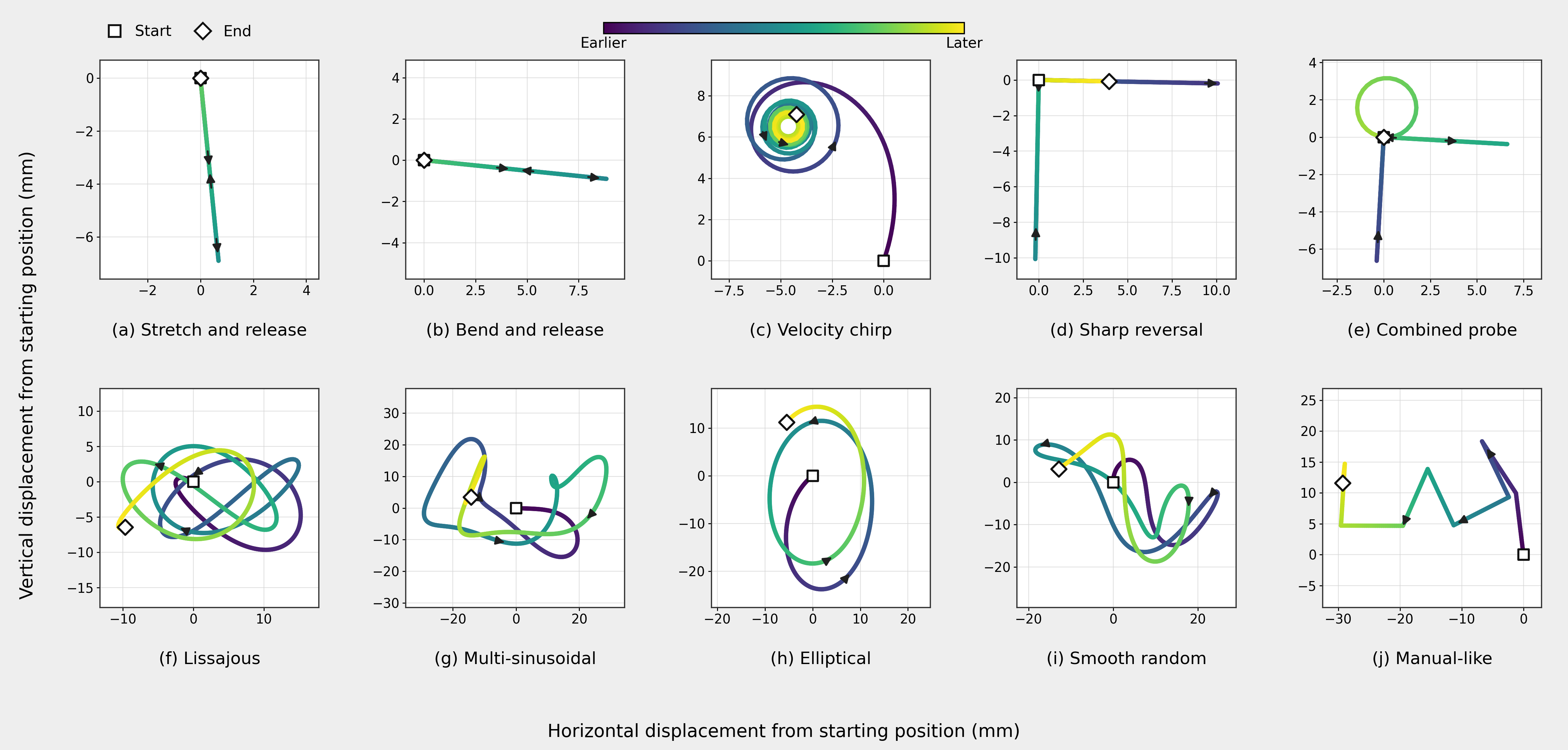}
    \caption{Different needle trajectories (thread tip) used to generate the training dataset.}
    \label{fig:motions}
\end{figure}

Ten informative needle motions are simulated for every virtual material:
stretch-and-release, bend-and-release, velocity chirp, sharp reversal,
combined probing, Lissajous, multi-sinusoidal, elliptical, smooth random, and manual-like motions. These different motions are represented in Fig.~\ref{fig:motions}.

Each trajectory contains 400 time steps with
\(\Delta t=0.048~\mathrm{s}\). The data is divided by virtual material, into 140 training materials, 30 validation
materials, and 30 test materials.

\subsubsection{Supervised Parameter Training}

Two trajectories generated using different motions but the same virtual
material are paired during training. Let
\(\widehat{\boldsymbol{\theta}}_{\mathrm{norm}}^{(a)}\) and
\(\widehat{\boldsymbol{\theta}}_{\mathrm{norm}}^{(b)}\) denote the predictions from two motions of the
same material, and let \(\boldsymbol{\theta}_{\mathrm{norm}}\) denote the normalized ground-truth vector of \eqref{eq:real_theta}.

The supervised parameter loss is
\begin{equation}
\mathcal{L}_{\mathrm{param}}
=
\left\|
\widehat{\boldsymbol{\theta}}_{\mathrm{norm}}^{(a)}
-
\boldsymbol{\theta}_{\mathrm{norm}}
\right\|_2^2
\end{equation}
A cross-motion consistency loss ensures the model identifies the material-dependent parameters rather than the particular excitation
motion:
\begin{equation}
\mathcal{L}_{\mathrm{cons}}
=
\left\|
\widehat{\boldsymbol{\theta}}_{\mathrm{norm}}^{(a)}
-
\widehat{\boldsymbol{\theta}}_{\mathrm{norm}}^{(b)}
\right\|_2^2
\end{equation}
The supervised training objective is \(\mathcal{L}_{\mathrm{sup}}=\mathcal{L}_{\mathrm{param}}+\lambda_{\mathrm{cons}}\mathcal{L}_{\mathrm{cons}}\).

Training used AdamW with a learning rate of \(3\times10^{-4}\), weight decay
\(10^{-4}\), batch size eight, and gradient clipping at one. A learning-rate scheduler
reduced the learning rate when the validation loss stopped improving, and the
checkpoint with the smallest validation loss was retained.

\subsubsection{Differentiable Multi-Step QP Fine-Tuning}

Training the network using only a parameter-level loss does not ensure that the predicted gains will produce a low tracking error during simulation.
The network is therefore fine-tuned using the trajectories generated by its predicted parameters. We initially included the loss \eqref{eq:tracking_loss} below during the training stage; however, this approach was computationally and memory intensive and produced
unstable gradients. Therefore, it
was introduced during this stage. For each training window beginning at \(t_0\), the propagation
in~\eqref{eq:dynamical_propagation} is initialized with
\(\widehat{\mathbf{z}}_{t_0}=\widetilde{\mathbf{z}}_{t_0}\) and evaluated for
\(H=12\) time steps using \(\widehat{\boldsymbol{\theta}}\) and the observed
needle inputs \(\widetilde{\mathbf{u}}_{t_0:t_0+H-1}\).

At each rollout step \(r\), the normalized tracking error is
\begin{equation}
    e_r
    =
    \frac{1}{N\Delta}
    \sum_{i=1}^{N}
    \left\|
        \widehat{\mathbf{x}}_{i,t_0+r}
        -
        \widetilde{\mathbf{z}}_{i,t_0+r}
    \right\|_2.
\end{equation}
Thus, the tracking loss is the average error over the rollout:
\begin{equation}
    \mathcal{L}_{\mathrm{track}}
    =
    \frac{1}{H}\sum_{r=1}^{H}e_r
    \label{eq:tracking_loss}
\end{equation}

The complete fine-tuning loss is
\begin{equation}
    \mathcal{L}_{\mathrm{rollout}}
    =
    \mathcal{L}_{\mathrm{track}}
    +
    0.2\mathcal{L}_{\mathrm{param}}
    +
    0.05\mathcal{L}_{\mathrm{cons}},
\end{equation}

Because the CBF--CLF--QP simulator is differentiable, the trajectory error can be
propagated through all 12 simulation steps to update the network. Each epoch used 48 randomly selected rollout windows, a batch size of
two, and a learning rate of \(10^{-5}\).

\subsubsection{Video-Level Parameter Inference}

To obtain a single parameter vector from a long video, the observed thread
motion is analyzed through multiple overlapping temporal windows. Each window
contains at most 400 frames, consecutive windows begin 200 frames apart, and
every second frame is supplied to the network. The windows capture
complementary portions of the thread response. The resulting window-level estimates
\(\widehat{\boldsymbol{\theta}}^{(k)}\), \(k=1,\ldots,K\), are averaged in the normalized parameter space to combine information across the complete video. The aggregated estimate is then mapped to the physical parameter ranges to obtain the final video-level parameter vector
\(\widehat{\boldsymbol{\theta}}\). It is assumed
that the thread material and its corresponding model parameters remain
constant throughout the video.

\section{Results}
\label{sec:results}

\begin{table}[t]
\centering
\caption{Parameter identification and QP replay for the asymmetric
flower-petal trajectory. Errors are mean node-trajectory errors.}
\label{tab:flower_petal_results}
\footnotesize
\setlength{\tabcolsep}{2pt}
\setlength{\arrayrulewidth}{0.35pt}
\renewcommand{\arraystretch}{1.12}

\begin{tabular}{
|>{\centering\arraybackslash}m{0.12\columnwidth}
|>{\centering\arraybackslash}m{0.28\columnwidth}
|>{\centering\arraybackslash}m{0.11\columnwidth}
|>{\centering\arraybackslash}m{0.11\columnwidth}
|>{\centering\arraybackslash}m{0.11\columnwidth}
|>{\centering\arraybackslash}m{0.16\columnwidth}|}
\Xhline{0.8pt}
\textbf{Material} &
\textbf{Estimate} &
$\boldsymbol{\gamma}$ &
$\boldsymbol{\alpha}$ &
$\boldsymbol{\kappa}$ &
\makecell{\textbf{QP error}\\\textbf{(mm)}} \\
\Xhline{0.8pt}

\multirow{3}{*}{1}
 & Ground Truth    & 11.174 & 13.561 & 0.642 & --    \\
\cline{2-6}
 & No Segmentation & 10.460 & 15.633 & 0.507 & 0.060 \\
\cline{2-6}
 & U-Net           &  2.489 &  2.267 & 0.526 & 0.639 \\
\Xhline{0.8pt}

\multirow{3}{*}{2}
 & Ground Truth    &  9.896 & 1.058 & 0.623 & --    \\
\cline{2-6}
 & No Segmentation & 15.757 & 1.815 & 0.433 & 0.043 \\
\cline{2-6}
 & U-Net           &  1.105 & 0.832 & 0.681 & 0.095 \\
\Xhline{0.8pt}

\multirow{3}{*}{3}
 & Ground Truth    & 13.695 & 1.628 & 0.838 & --    \\
\cline{2-6}
 & No Segmentation & 19.845 & 2.292 & 0.515 & 0.257 \\
\cline{2-6}
 & U-Net           &  1.940 & 0.967 & 0.725 & 0.346 \\
\Xhline{0.8pt}

\multirow{3}{*}{4}
 & Ground Truth    & 17.564 & 0.561 & 0.737 & --    \\
\cline{2-6}
 & No Segmentation & 14.808 & 1.454 & 0.408 & 0.080 \\
\cline{2-6}
 & U-Net           &  2.562 & 1.660 & 0.323 & 0.123 \\
\Xhline{0.8pt}

\multirow{3}{*}{5}
 & Ground Truth    & 18.432 & 4.119 & 0.514 & --    \\
\cline{2-6}
 & No Segmentation & 24.500 & 5.503 & 0.305 & 0.023 \\
\cline{2-6}
 & U-Net           &  1.340 & 0.936 & 0.758 & 0.274 \\
\Xhline{0.8pt}

\multirow{3}{*}{6}
 & Ground Truth    & 10.000 & 1.000 & 0.800 & --    \\
\cline{2-6}
 & No Segmentation & 15.469 & 1.752 & 0.458 & 0.177 \\
\cline{2-6}
 & U-Net           &  0.755 & 0.594 & 0.803 & 0.194 \\
\Xhline{0.8pt}
\end{tabular}

\end{table}

\begin{figure}
    \centering
    \includegraphics[width=0.8\linewidth, height=0.4\textheight, keepaspectratio]{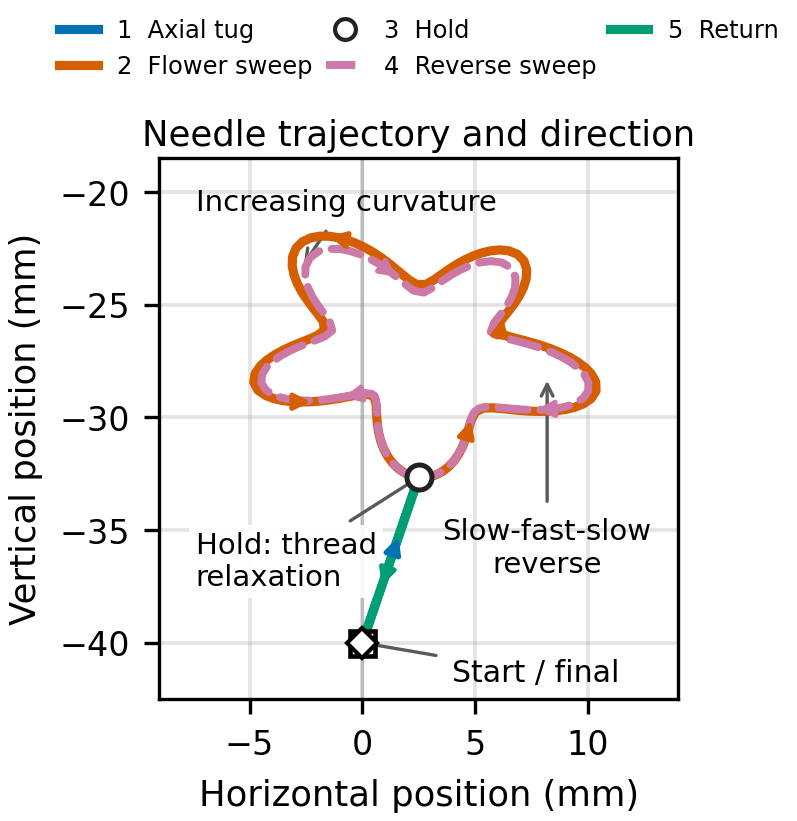}
    \caption{Asymmetric-flower-petal needle trajectory used for evaluation, comprising a forward flower-petal sweep, a variable-speed reverse, and a return to the initial position.}
    \label{fig:infinity}
\end{figure}

\begin{figure}
    \centering
    \includegraphics[width=0.8\linewidth,
    height=0.4\textheight,
    keepaspectratio]{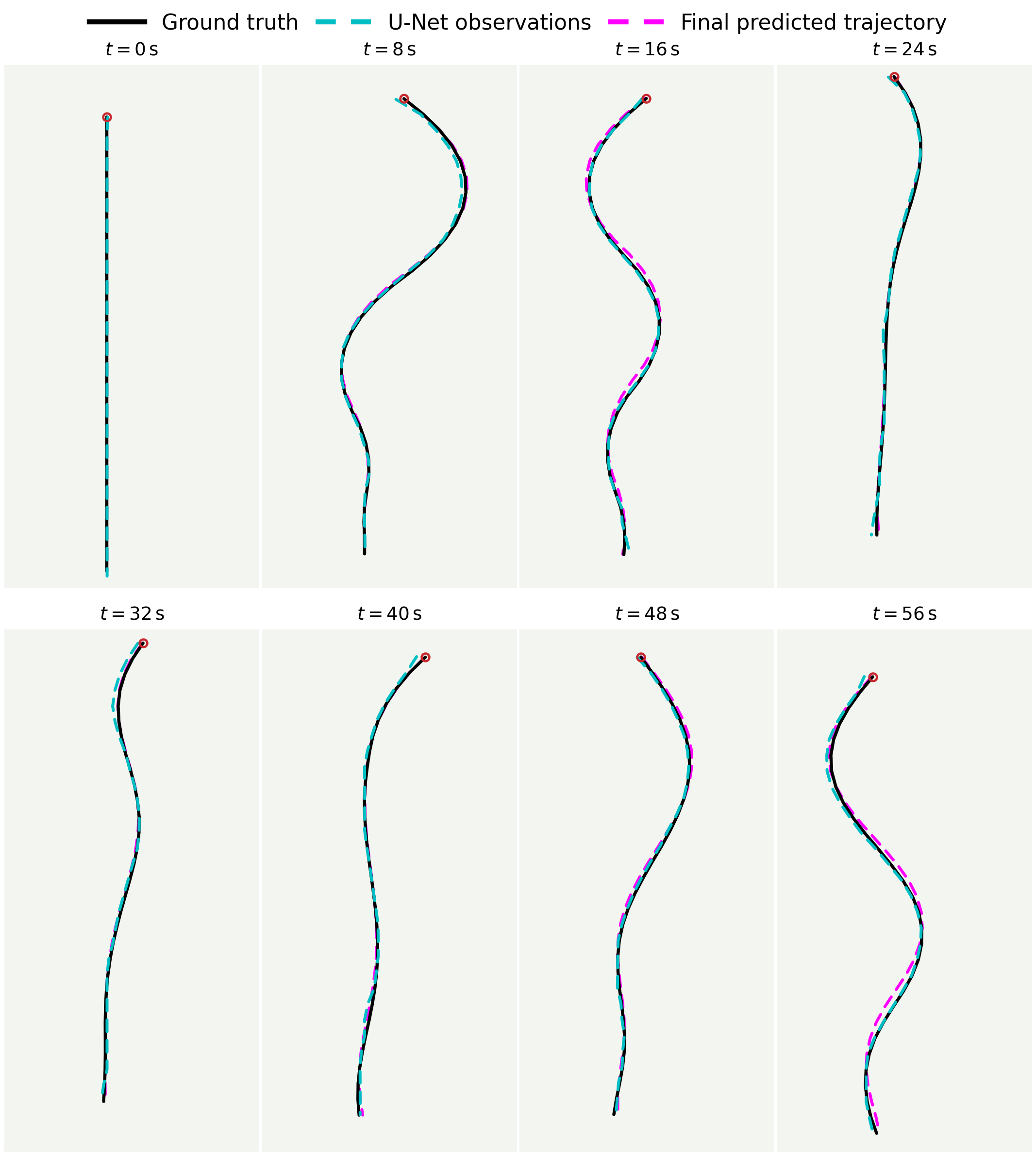}
    \caption{Representative thread configurations during the asymmetric flower-petal motion experiment. Black denotes the ground-truth trajectory, cyan denotes the U-Net observations, and magenta denotes the trajectory predicted by our model.}
    \label{fig:inference}
\end{figure}

\subsubsection{Reference Model and Evaluation Motion}

The proposed video-to-model framework was evaluated using a new set of virtual materials not included in the training set.
The parameter identification was performed with and without the U-Net segmentation network to assess segmentation accuracy. The QP modeling  errors are reported in Table~\ref{tab:flower_petal_results}.

An asymmetric flower-petal needle motion was chosen for testing because: i) it isn't a part of the the training set motions and ii) provides sufficient complexity to effectively excite the thread to validate our model's parameter identification. The motion is shown in Fig.~\ref{fig:infinity}.

For each virtual material, the resulting sequence contained \(1240\) frames sampled
at intervals of \(0.048\,\mathrm{s}\), corresponding to a frame rate of
approximately \(20.83\,\mathrm{Hz}\) and a total duration of
\(59.52\,\mathrm{s}\).

\subsubsection{Thread Reconstruction from Video}

Ordered centerlines were successfully obtained in most of the frames, and had a corresponding trace success rate of \(99.92\%\). Occasional segmentation artifacts and tracing errors displaced individual nodes, introducing temporal jitter and inaccurate velocity estimates. 

\subsubsection{Automatic Model Configuration}

The spatio-temporal CNN estimated the model parameters from centerlines extracted by the U-Net segmentation network. When evaluated independently using ground-truth centerlines, the model identified near-correct parameters that produced very low QP replay errors, demonstrating its parameter-identification capability. Although U-Net accurately segmented the thread, in some instances, temporal noise in its centerline degraded the predicted parameters, indicating that further fine-tuning of the network is required to identify parameters accurately. Despite mismatch in parameters, the resulting tracking error remains low and comparable in many of the cases.

\subsubsection{Trajectory Reconstruction}

The predicted gains were inserted into the CBF--CLF--QP model. The simulation
was initialized using the first reconstructed thread configuration and driven
using the needle motion recovered from the video. The resulting trajectory was
compared with the exact simulator
trajectory generated using ground truth \(\boldsymbol{\theta}\).

Fig.~\ref{fig:inference} shows snapshots of the simulated silk thread (virtual material 6 in Table~\ref{tab:flower_petal_results}) in different configurations as the needle follows an asymmetric flower-petal trajectory.
The predicted model reproduced the simulated silk reference trajectory with a mean
node error of \(0.194\,\mathrm{mm}\).

\section{Conclusion}
\label{sec:conclusion}

This work presented a video-to-model framework for automatically configuring a given CBF--CLF--QP model of a suture thread. The proposed framework first localizes the thread in video and reconstructs its ordered node trajectory.
A spatio-temporal CNN then uses the observed thread and needle motion to
estimate a single set of model parameters, which is subsequently used to
simulate the thread dynamics. Experiments  demonstrated that the configured model can reproduce the behavior of the
observed thread on a held-out trajectory with very low error without requiring manual
parameter tuning.

The results also revealed two important limitations. First, occasional segmentation and centerline-tracing errors introduced inaccurate node observations, leading to errors in the parameter predictions. Second, some excitation trajectories did not sufficiently distinguish the effects of individual parameters, making them difficult to identify uniquely. Future work will therefore focus on more robust visual tracking, parameter-sensitive excitation trajectories, and experiments with real suture threads made from different materials.






\end{document}